\documentclass[10pt,twocolumn,letterpaper]{article}

\usepackage{iccv}
\usepackage{times}
\usepackage{epsfig}
\usepackage{graphicx}
\usepackage{amsmath}
\usepackage{amssymb}
\usepackage{enumitem}
\usepackage[accsupp]{axessibility}  

\setlist{nolistsep}

\DeclareMathOperator{\softmax}{\operatorname{softmax}}
\DeclareMathOperator{\relu}{\operatorname{ReLU}}

\usepackage[pagebackref=true,breaklinks=true,letterpaper=true,colorlinks,bookmarks=false]{hyperref}

\iccvfinalcopy 

\def\iccvPaperID{11} 

\ificcvfinal\fi

\begin{document}

\title{Graph CNN for Moving Object Detection in Complex Environments from Unseen Videos}

\author{Jhony H. Giraldo$^1$, Sajid Javed$^2$, Naoufel Werghi$^2$, Thierry Bouwmans$^1$\\
$^1$La Rochelle Université, $^2$Khalifa University of Science and Technology\\
{\tt\small \{jgiral01, tbouwman\}@univ-lr.fr, \{sajid.javed, naoufel.werghi\}@ku.ac.ae}}

\maketitle
\ificcvfinal\thispagestyle{empty}\fi

\begin{abstract}
   Moving Object Detection (MOD) is a fundamental step for many computer vision applications. 
   MOD becomes very challenging when a video sequence captured from a static or moving camera suffers from the challenges: camouflage, shadow, dynamic backgrounds, and lighting variations, to name a few.
   Deep learning methods have been successfully applied to address MOD with competitive performance.
   However, in order to handle the overfitting problem, deep learning methods require a large amount of labeled data which is a laborious task as exhaustive annotations are always not available.
   Moreover, some MOD deep learning methods show performance degradation in the presence of unseen video sequences because the testing and training splits of the same sequences are involved during the network learning process.
   In this work, we pose the problem of MOD as a node classification problem using Graph Convolutional Neural Networks (GCNNs).
   Our algorithm, dubbed as GraphMOD-Net, encompasses instance segmentation, background initialization, feature extraction, and graph construction.
   GraphMOD-Net is tested on unseen videos and outperforms state-of-the-art methods in unsupervised, semi-supervised, and supervised learning in several challenges of the Change Detection 2014 (CDNet2014) and UCSD background subtraction datasets.
\end{abstract}

\section{Introduction}
\label{sec:introduction}

Moving Object Detection (MOD) is a crucial problem in computer vision for applications such as video surveillance, pose estimation, and intelligent visual observation of animals, to name just a few \cite{giraldo2019camera,garcia2020background}.
MOD mainly aims to separate the moving objects known as foreground from the static component known as background \cite{bouwmans2014traditional}. 
In the literature, MOD has been considered as a binary classification problem where each pixel is predicted for either background or foreground component in a sequence taken from a static or moving camera. Therefore, this problem is also known as background-foreground segregation \cite{bouwmans2017decomposition}.
MOD becomes a very challenging problem because of the presence of dynamic variations in the background scene such as illumination variations, swaying bushes, camouflage, intermittent object motion, shadows, and jittering effects, to name a few \cite{wang2014cdnet}.
In addition, videos that are taken from PanTiltZoom (PTZ) and moving cameras also pose more challenges for efficient MOD.
Several methods have been proposed to improve the performance of MOD under challenging scenarios such as those presented in \cite{oreifej2012simultaneous,pham2014detection,giraldo2020graph}.
However, there is no unique unsupervised or supervised method that can effectively handle all the aforementioned challenges in real scenarios \cite{tezcan2020bsuv,wang2014cdnet}.

\begin{figure*}
    \centering
    \includegraphics[width=0.86\textwidth]{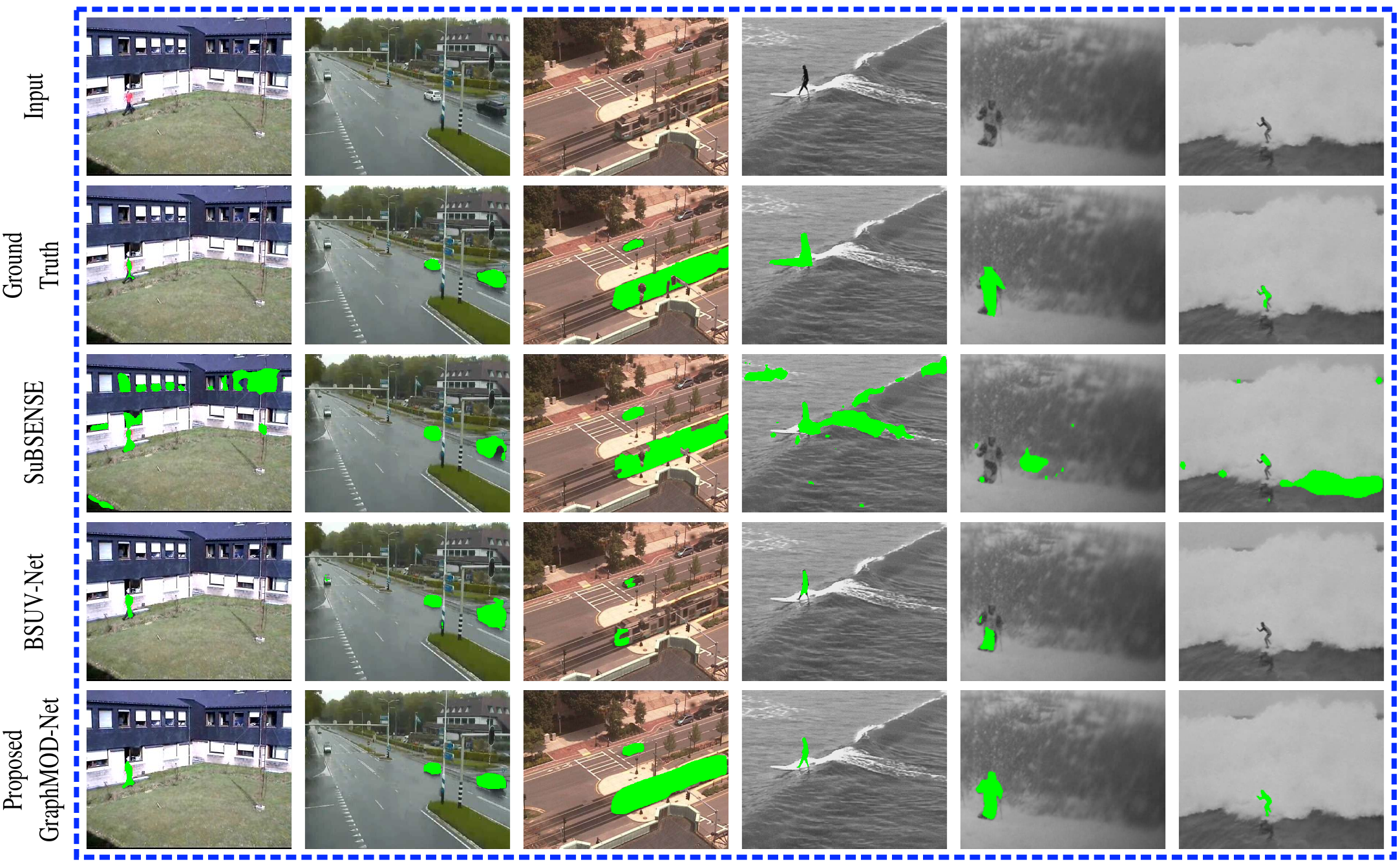}
    \caption{Comparisons of the visual results of the proposed GraphMOD-Net algorithm with two state-of-the-art methods for MOD on six very challenging video sequences taken from CDNet2014 \cite{wang2014cdnet} and UCSD \cite{mahadevan2009spatiotemporal} datasets, including moving camera sequences plus dynamic backgrounds. From left to right: zoom in zoom out, two-position PTZ cameras, intermittent pan, surfers, skiing, and surf sequences. From top to bottom: input frame, the ground truth of MOD, SuBSENSE \cite{st2014subsense}, BSUV-Net \cite{tezcan2020bsuv}, and our proposed GraphMOD-Net.}
    \label{fig:introductory_image}
\end{figure*}

Many state-of-the-art MOD methods such as SuBSENSE \cite{st2014subsense} work well for sequences taken from a static camera; however, they show performance degradation in the presence of moving camera sequences as shown in Fig. \ref{fig:introductory_image}. And in fact, none of the methods proposed so far can address all challenges posed by static as well as moving cameras.
In the current work, we contribute to bridging these gaps by formulating the MOD problem on Graph Convolutional Neural Networks (GCNNs).
We propose a novel semi-supervised algorithm dubbed Graph MOD Network (GraphMOD-Net) based on GCNNs \cite{kipf2017semi}. 
GraphMOD-Net models the instances in videos as nodes embedded in a graph. 
The instances are obtained with a Mask Region-CNN (Mask R-CNN) \cite{he2017mask} or Cascade Mask R-CNN \cite{cai2019cascade}. 
The representation of the nodes is obtained with background initialization, optical flow, intensity, and texture features \cite{lucas1981iterative,ojala2002multiresolution}. 
Moreover, these nodes are associated either with the class foreground or background using ground-truth information and finally, a GCNN is trained with a small percentage of labeled nodes to perform a semi-supervised learning classification \cite{kipf2017semi} with an unseen scheme \cite{tezcan2020bsuv}.
GraphMOD-Net outperforms state-of-the-art methods in several challenges of the Change Detection 2014 (CDNet2014) \cite{wang2014cdnet} and UCSD background subtraction \cite{mahadevan2009spatiotemporal} datasets. The main contributions of this paper are summarized as follows:
\begin{itemize}[leftmargin=*]
\itemsep0em
\item We pose the problem of MOD as a binary classification problem on the graph where each node is classified into two distinct components including background and foreground using GCNNs.
\item We perform rigorous experiments using the CDNet2014 dataset \cite{wang2014cdnet} for MOD. Our results demonstrate that GraphMOD-Net uses a limited amount of labeled data and outperforms some state-of-the-art classical and deep learning methods.
\end{itemize}

The rest of the paper is organized as follows. Section \ref{sec:related_works} presents the related works. Section \ref{sec:algorithm} explains the basic concepts and the proposed GraphMOD-Net. Section \ref{sec:experimental_framework} introduces the experimental framework. And finally, Sections \ref{sec:results_discussion} and \ref{sec:conclusions} present the results and conclusions, respectively.

\section{Related Work}
\label{sec:related_works}

The MOD methods can be categorized into unsupervised, semi-supervised, and supervised learning methods. The unsupervised methods can be classified as statistical \cite{bouwmans2008background,narayana2014background,zuo2019moving}, subspace learning \cite{bouwmans2009subspace}, and robust principal component analysis models \cite{bouwmans2014robust,rodriguez2016incremental}. These methods do not successfully adapt to complex scenarios in MOD. Inspired by the success of deep CNNs on a wide variety of visual recognition tasks \cite{russakovsky2015imagenet}, several studies have also been proposed to handle the MOD problem in a fully supervised manner \cite{garcia2019foreground,bouwmans2019deep}. However, these deep learning methods usually require a large amount of training data. The annotation of such data is an intensive task and can be very expensive to obtain the required degree of quality.
Even though some works have tried to explain the success of deep learning methods for MOD \cite{minematsu2018analytics}, there is no concrete evidence in the literature about the sample complexity required in the deep learning regimen \cite{du2018many}. 
Furthermore, being trained and tested on sequences derived from the same videos, most of the deep learning methods for MOD do not show good generalization to unseen videos exhibiting unpredictable changes. The performance degradation was confirmed in CDNet2014 (FgSegNetv2 \cite{lim2019learning}) for which the performance of leading methods dropped dramatically when evaluated on unseen videos as reported in \cite{tezcan2020bsuv}.
For a complete review of supervised and unsupervised methods, we invite the readers to the survey papers \cite{bouwmans2008background,bouwmans2014traditional,bouwmans2014robust,bouwmans2019deep}.

\begin{figure*}
    \centering
    \includegraphics[width=0.89\textwidth]{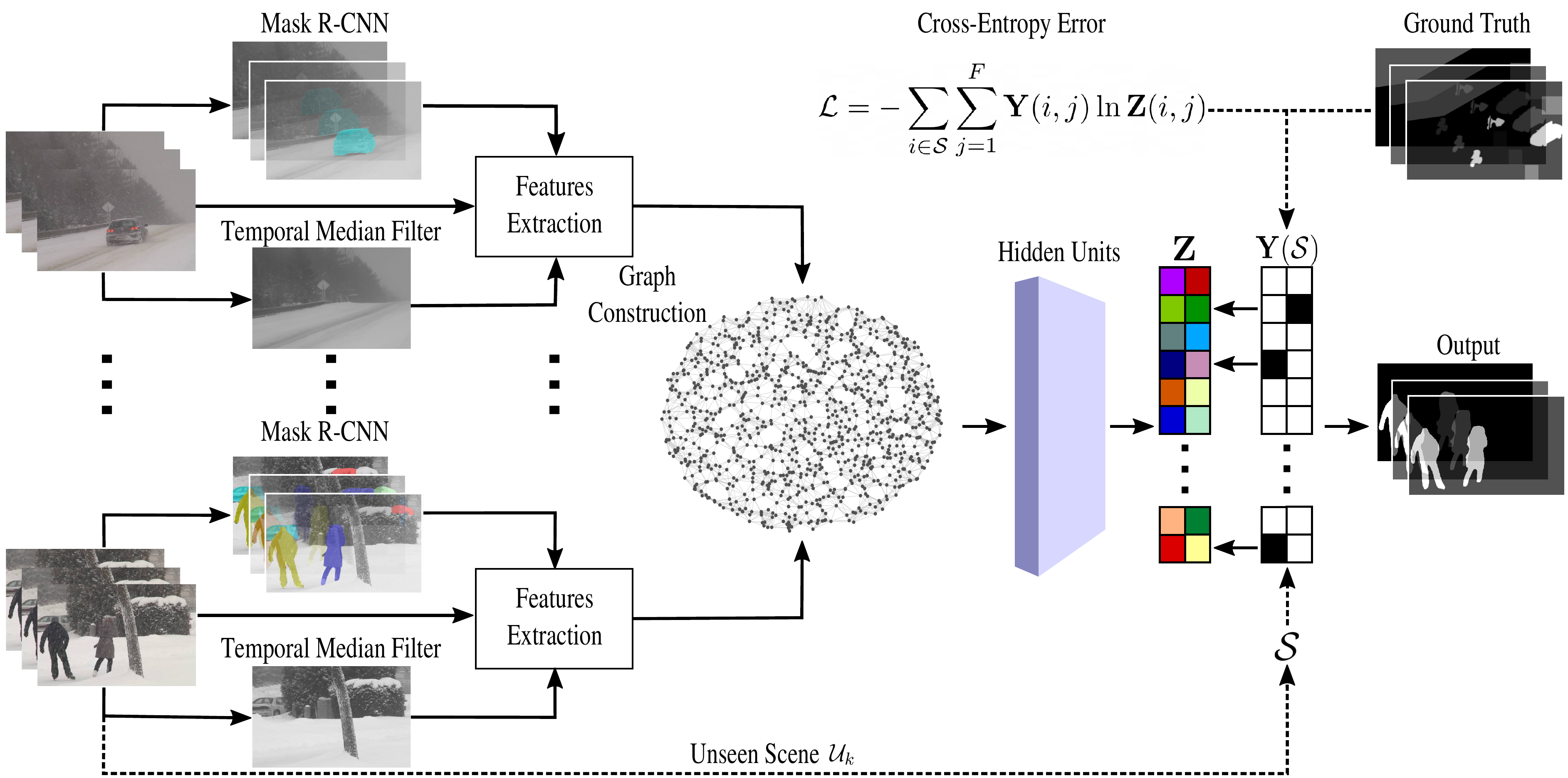}
    \caption{GraphMOD-Net uses background initialization and instance segmentation. Each instance represents a node in a graph using motion, intensity, and texture features. Finally, a GCNN classifies if each node is a moving or static object with an unseen scheme.}
    \label{fig:pipeline}
\end{figure*}

Giraldo and Bouwmans \cite{giraldo2020graphbgs} have recently proposed a Graph-based semi-supervised learning method for BackGround Subtraction (GraphBGS). Leveraging the theory of sampling and graph signal reconstruction, this framework found applications in MOD \cite{ortega2018graph}.
GraphBGS exploits a variational approach to solve the semi-supervised learning problem \cite{pesenson2009variational}, assuming that the underlying signals corresponding to the background/foreground nodes are smooth in the graph \cite{chen2015discrete}. Therefore, this method shows performance degradation if the smooth prior assumption is not fully satisfied. In this work, we propose a semi-supervised learning algorithm dubbed as GraphMOD-Net. Our algorithm extends the work of Giraldo and Bouwmans \cite{giraldo2020graphbgs} by increasing modeling capacity and avoiding the explicit smoothness assumption with GCNNs. GraphMOD-Net has the advantage of requiring less training data than deep learning methods while adapting to complex MOD scenarios, unlike unsupervised methods.


\section{Proposed Algorithm}
\label{sec:algorithm}


Our proposed algorithm consists of several components including instance segmentation, background model initialization, features extraction, graph construction, and GCNN training, as shown in the block diagram in Fig. \ref{fig:pipeline}.
In the first step, we compute the instance segmentation mask of the input sequence using the Mask R-CNN method \cite{he2017mask}.
The instance segmentation mask assists our algorithm to get the prior knowledge of the object instances in the training videos.
We also compute the initial background model using the temporal median filtering method.
In the second step, the instances and background model are used to derive the features to be used in the graph construction.
We use texture features, motion features using optical flow estimation \cite{lucas1981iterative}, and intensity features. However, other types of features can be used to represent the nodes in the graph \cite{giraldo2020graph}.
In the second step, we construct the graph where a node is assigned to each object instance represented by the group of the aforementioned features. With the so obtained graph representation, we train a GCNN to classify the nodes into either background or foreground. The different components of the algorithm will be described next.

\subsection{Instance Segmentation Description}

We used the Mask R-CNN \cite{he2017mask}, with a Residual Network of 50 layers (ResNet50) \cite{he2017mask} and Feature Pyramid Network (FPN) \cite{lin2017feature} as the backbone, for instance segmentation in the CDNet2014. While for the UCSD dataset we used Cascade Mask R-CNN \cite{cai2019cascade} with ResNeSt of 200 layers \cite{zhang2020resnest}.
Both instance segmentation networks are trained with the Common Objects in Context (COCO) 2017 dataset \cite{lin2014microsoft}. Each output (mask) of the instance segmentation method represents a node in our graph.
We note that super-pixels can be used as an alternative for object instances in GraphMOD-Net.
However, super-pixels and pixel-based approaches, in general, produce a huge-sized graph, making the computation cost-prohibitive and the implementation impractical \cite{parada2019blue,lu2019fast}.


\subsection{Graph Construction}

A graph $G$ can be represented with two sets as $(\mathcal{V,E})$, where $\mathcal{V}=\{1,\dots,N\}$ is the set of nodes, with cardinality $\vert \mathcal{V} \vert = N$ is the number of nodes of $G$. $\mathcal{E}=\{(i,j)\}$ is the set of edges, where $(i,j)$ represents an edge between the nodes $i$ and $j$. The adjacency matrix of $G$ is $\mathbf{A} \in \mathbb{R}^{N \times N}$, where $\mathbf{A}(i,j)$ is the weight of the edge $(i,j)$. A graph is called unweighted when $\mathbf{A}(i,j)=1~\forall~(i,j) \in \mathcal{E}$. For undirected graphs $\mathbf{A}$ is symmetric, \ie, the weights between the edges $(i,j)$ and $(j,i)$ are the same. Finally, $\mathbf{D} \in \mathbb{R}^{N \times N}$ is the degree matrix of $G$ defined as a diagonal matrix such that $\mathbf{D}(i,i)=\sum_{j=1}^N \mathbf{A}(i,j)~\forall~i \in \mathcal{V}$.

In our graph, each node is represented by a $853$-dimensional vector $\mathbf{x}_v$ for $v \in \mathcal{V}$, and grouped in the matrix $\mathbf{X}=[\mathbf{x}_1,\dots,\mathbf{x}_N]^{\mathsf{T}}$. The feature vector $\mathbf{x}_v$ is obtained using a temporal median filter for background initialization, optical flow, intensity, and texture features, after converting the videos to gray-scale. Let $\mathbf{I}_v^t$ be the Region of Interest (RoI) image of node $v \in \mathcal{V}$ in the current ($t$) frame. The RoI is the group of pixels belonging to an object instance produced by the instance segmentation method (the bounding box output of Mask R-CNN). Let $\mathcal{P}_v$ be the set of pixel-indices corresponding to the mask object $v$ (the mask output of Mask R-CNN). Let $\mathbf{B}$ be the output image of the temporal median filter and let $\mathbf{B}_v$ be the RoI image of the background of node $v$, \ie, we are extracting the bounding box of $v$ in $\mathbf{B}$.
We compute the optical flow vectors, of the current frame with support in $\mathcal{P}_v$, and their amplitudes and orientations. Afterward, we construct histograms and their descriptive statistics: minimum, maximum, mean, standard deviation, mean absolute deviation, and range. We also use 
the RoI images $\mathbf{I}_v^t$, $\mathbf{I}_v^{t-1}$, $\mathbf{B}_v$ and $\vert \mathbf{I}_v^t - \mathbf{B}_v \vert$ to derive a texture representation, with local binary patterns \cite{ojala2002multiresolution}, and the intensity histograms. All the previous features are concatenated into thee feature vector $\mathbf{x}_v$ 
representing the node $v$.


The construction of the undirected graph $G$ is performed using the k-nearest neighbors algorithm with $\text{k}=30$ in the matrix $\mathbf{X}$. The elements of the adjacency matrix are computed with a Gaussian kernel $\mathbf{A}(i,j) = \exp{({-\frac{d(i,j)^2}{\rho^2}})}$, where $d(i,j)=\Vert \mathbf{x}_i - \mathbf{x}_j \Vert$, and $\rho$ is the standard deviation of the Gaussian function, set to $\rho = \frac{1}{\vert \mathcal{E} \vert + N}\sum_{(i,j) \in \mathcal{E}} d(i,j)$. Furthermore, the full set of labels $\mathbf{Y}\in \mathbb{R}^{N\times F}$ of our algorithm is a matrix encoding the classes of all nodes in $\mathcal{V}$, where $F=2$ (background and foreground). Each row of $\mathbf{Y}$ is represented with the so-called one-hot vector: background $[1,0]$, and foreground $[0,1]$. 


\subsection{Graph Semi-Supervised Learning Algorithm}

The previous semi-supervised algorithm GraphBGS \cite{giraldo2020graphbgs} relies on the sampling and reconstruction of graph signals \cite{chen2015discrete,ortega2018graph,anis2018sampling,parada2019blue}. As a consequence, Giraldo and Bouwmans \cite{giraldo2020graphbgs} defined the sampled nodes (or training set in our case) as a subset of nodes $\mathcal{S} \subset \mathcal{V}$ with $\mathcal{S}=\{s_1,s_2,\dots,s_m\}$, where $m = \vert \mathcal{S}\vert \leq N$ is the number of sampled nodes. GraphBGS applied a variational method in graphs \cite{pesenson2009variational} to solve the semi-supervised learning problem, the application of this variational method assumes that the background/foreground nodes are smooth in the graph. Unlike GraphBGS, our algorithm avoids the smoothness assumption by solving the learning problem with GCNNs, and we refer to the set $\mathcal{S}$ as the training set in GraphMOD-Net.

The layer-wise propagation rule of our GraphMOD-Net, inspired from Kipf and Welling \cite{kipf2017semi} method, is given as follows:
\begin{equation}
    \mathbf{H}^{(l+1)}=\sigma(\tilde{\mathbf{D}}^{-\frac{1}{2}}\tilde{\mathbf{A}}\tilde{\mathbf{D}}^{-\frac{1}{2}}\mathbf{H}^{(l)}\mathbf{W}^{(l)})
    \label{eqn:propagation_rule}
\end{equation}
where $\tilde{\mathbf{A}}=\mathbf{A}+\mathbf{I}$ is the adjacency matrix of $G$ with added self-connections\footnote{The self-connections given by $\tilde{\mathbf{A}}=\mathbf{A}+\mathbf{I}$ were introduced by Kipf and Welling \cite{kipf2017semi} to avoid numerical instabilities and exploding/vanishing gradients.}, $\mathbf{I}$ is the identity matrix, $\tilde{\mathbf{D}}$ is the degree matrix of $\tilde{\mathbf{A}}$, $\mathbf{W}^{(l)}$ is the matrix of trainable weights in layer $l$, $\sigma(\cdot)$ denotes an activation function, and $\mathbf{H}^{(l)}$ is the matrix of activations in layer $l$ such that $\mathbf{H}^{(0)}=\mathbf{X}$ is the input matrix denoting the representation of the nodes. The propagation rule in Eqn. (\ref{eqn:propagation_rule}) is motivated by the first-order approximation of localized spectral filters on graphs \cite{hammond2011wavelets,defferrard2016convolutional}. For further details, the reader is referred to Kipf and Welling paper \cite{kipf2017semi}.

GraphMOD-Net uses a GCNN with one hidden layer to solve the semi-supervised learning problem. The forward of our model is computed using the propagation rule of Eqn. (\ref{eqn:propagation_rule}) as follows:
\begin{align}
    \nonumber
    \mathbf{Z}=f(\mathbf{X,A}) = \softmax \left(\tilde{\mathbf{D}}^{-\frac{1}{2}}\tilde{\mathbf{A}}\tilde{\mathbf{D}}^{-\frac{1}{2}} \right. & \\
    \label{eqn:forward_gcnn}
    \left. \relu\left(\tilde{\mathbf{D}}^{-\frac{1}{2}}\tilde{\mathbf{A}}\tilde{\mathbf{D}}^{-\frac{1}{2}} \mathbf{XW}^{(0)}\right)\mathbf{W}^{(1)}\right) &,
\end{align}
where $\mathbf{Z}$ is the output of the GCNN; $\mathbf{W}^{(0)} \in \mathbb{R}^{C\times H}$ is the weights of the first hidden layer such that $C$ is the dimension of vector $\mathbf{x}_v$, \ie, $C=853$, and $H$ is the number of feature maps; $\mathbf{W}^{(1)} \in \mathbb{R}^{H\times F}$ is the weights of the output layer such that $F$ is the number of classes (\ie, $F=2$: background and foreground); $\relu(\cdot)=\max(0,\cdot)$ is the rectified linear unit; and finally the softmax activation function is defined as $\softmax(x_i)=\frac{1}{S}\exp{(x_i)}$ where $S=\sum_i \exp{(x_i)}$. Here the softmax function is applied row-wise. We set the number of feature maps $H$ in the hidden layer to $16$ in our model. The loss function of this GCNN is defined as the cross-entropy error over all labeled nodes:
\begin{equation}
    \mathcal{L}=-\sum_{i \in \mathcal{S}} \sum_{j=1}^F \mathbf{Y}(i,j) \ln{\mathbf{Z}(i,j)}.
    \label{eqn:cross_entropy}
\end{equation}
We use the Adam optimizer \cite{kingma2015adam} for training and set the training batch to the full training set in each iteration (this is possible since the whole database fits in memory).

\section{Experimentation}
\label{sec:experimental_framework}

This section introduces the databases used in this paper, the experimentation protocol, the evaluation metrics, the experiments, and the implementation details of GraphMOD-Net.

\subsection{Databases}

We use the Change Detection 2014 (CDNet2014) \cite{wang2014cdnet} and the UCSD background subtraction \cite{mahadevan2009spatiotemporal} datasets in this work. CDNet2014 contains $11$ challenges including, bad weather, low frame rate, night videos, PTZ, turbulence, baseline, dynamic background, camera jitter, intermittent object motion, shadow, and thermal. Each challenge has between $4$ to $6$ videos, and each video has from $600$ up to $7999$ images. Certain frames in each video contain ground-truth images, consisting of pixel-level annotations of foreground and background. On the other hand, UCSD contains $18$ videos mainly composed of moving camera sequences, with $30$ up to $246$ frames each sequence. Each video of UCSD is partially or fully annotated with pixel-level ground-truth images of foreground and background.

\subsection{Experimentation protocol}
\label{sec:training_validation_test}

The selection of the training, validation, and test nodes is based on an agnostic-video evaluation methodology in order to best evaluate the performance of GraphMOD-Net on unseen videos.
To that end, the testing nodes should come from different videos with respect to the training nodes. Formally, let $\mathcal{V}_k$ be a partition of the whole set of nodes $\mathcal{V}_k \subset \mathcal{V}$, and let $\mathcal{U}_k \subset \mathcal{V}_k$ be the set of nodes corresponding to the unseen videos for that specific partition $k$. Then, the training set $\mathcal{S} \subset \mathcal{V}_k$ is randomly chosen from a subset of possible nodes $\mathcal{V}_k \setminus \mathcal{U}_k$. In other words, the training set is $\mathcal{S}$, the test set is $\mathcal{U}_k$, and the validation set $\mathcal{T}_k$ is randomly chosen from the subset $\mathcal{T}_k \subset \mathcal{V}_k \setminus (\mathcal{S} \bigcup \mathcal{U}_k)$.

For CDNet2014 we adopt the same partition of the database as Tezcan \etal \cite{tezcan2020bsuv}. The database is divided in 18 sets, \ie, $k=\{1,\dots,18\}$, where each split contains a list of training and test sequences. GraphMOD-Net randomly selects a small subset of nodes $\mathcal{S} \subset \mathcal{V}_k$ and a small set $\mathcal{T}_k$ for validation from the training sequences to train and then evaluate the GCNN on unseen videos. For example, the first partition is given such that $\mathcal{V}_1$ contains the nodes corresponding to the challenges: baseline, bad weather, intermittent object motion, low framerate, and shadow; while the set of unseen videos $\mathcal{U}_1$ contains the sequences: office, and PETS2006 from baseline; and backdoor, copy machine, and people in shade from shadow. The cardinality of $\mathcal{T}_k$ is fixed for all experiments such that $\vert \mathcal{T}_k \vert = N\times 0.01$, \ie, we are using $1\%$ of the dataset as validation. The 18 partitions for the CDNet2014 is given in the supplementary material of Tezcan \etal paper \cite{tezcan2020bsuv}.

For UCSD we train our GraphMOD-Net in a subset of nodes of CDNet2014 and we test in UCSD dataset. For the training and validation sets we randomly choose nodes from the challenges baseline, dynamic background, shadow, and PTZ of CDNet2014.

\subsection{Evaluation Metrics}

GraphMOD-Net is evaluated using the F-measure defined as follows:
\begin{gather}
    \label{eqn:f-measure}
    \text{F-measure} = 2\frac{\text{Precision}\times \text{Recall}}{\text{Precision}+\text{Recall}},\\
    \text{Recall} = \frac{\text{TP}}{\text{TP}+\text{FN}}, \text{ Precision} = \frac{\text{TP}}{\text{TP}+\text{FP}},
\end{gather}
with TP, FP, and FN the number of True Positives, False Positives, and False Negatives pixels, respectively.

\subsection{Experiments}

For CDNet2014, GraphMOD-Net constructs a graph for the whole dataset, resulting in an undirected graph of $258956$ nodes. For each partition of the dataset a percentage of the total amount of nodes in the set $\mathcal{M}=\{ 0.001, 0.005, 0.05, 0.1\}$ is used for training and then evaluated with unseen videos. For example, $\mathcal{S}$ contains $259$ nodes for $0.001 \in \mathcal{M}$ (keep in mind that usually each image contains several nodes as shown in Fig. \ref{fig:pipeline}). The training of our GCNN is validated with a Monte Carlo cross-validation with $3$ repetitions for each training density and split of the database. As a consequence, we train $18 \times 3 \times \vert \mathcal{M} \vert = 216$ GCNNs for the validation of GraphMOD-Net.

For UCSD, GraphMOD-Net constructs a graph with the whole UCSD dataset plus the challenges baseline, dynamic background, shadow, and PTZ of CDNet2014; resulting in a graph of $104261$ nodes. In this case, the test set is the whole UCSD dataset, and a small number of nodes in the set $\mathcal{M}$ is used for training. A Monte Carlo cross-validation with 5 repetitions is performed for each training density, \ie, we train $5 \times \vert \mathcal{M} \vert = 20$ GCNNs for UCSD dataset. For the sake of clarification, the metrics in the experiments are computed on the unseen set of nodes in each experiment. The CDNet2014 and UCSD experiments follow the guidelines for the training, validation, and testing sets of Section \ref{sec:training_validation_test}.

\begin{table*}
\centering
\caption{Comparison of the average F-measure over nine challenges of the CDNet2014 dataset, namely: bad weather (BWT), baseline (BSL), camera jitter (CJI), dynamic background (DBA), intermittent object motion (IOM), low frame rate (LFR), PTZ, shadow (SHW), and thermal (THL). The best and second best performing methods for each challenge are shown in {\color{red}\textbf{red}} and {\color{blue}\textbf{blue}}, respectively.}
\label{tbl:CDNet_results}
\begin{scriptsize}
\begin{tabular}{lcccccccccc}
\hline
\multicolumn{1}{l|}{Method} & BWT & BSL & CJI & DBA & IOM & LFR & PTZ & SHW & \multicolumn{1}{c|}{THL} & Overall \\
\hline
\multicolumn{1}{l|}{GraphCutDiff (unsupervised)} & {\color{red}\textbf{0.879}} & 0.715 & 0.549 & 0.539 & 0.402 & 0.513 & 0.372 & 0.723 & \multicolumn{1}{c|}{0.579} & 0.5857 \\
\multicolumn{1}{l|}{SuBSENSE (unsupervised)} & 0.862 & 0.950 & {\color{blue}\textbf{0.815}} & 0.818 & 0.657 & {\color{blue}\textbf{0.645}} & 0.348 & 0.899 & \multicolumn{1}{c|}{{\color{blue}\textbf{0.817}}} & 0.7567 \\
\multicolumn{1}{l|}{WisenetMD (unsupervised)} & 0.862 & 0.949 & {\color{red}\textbf{0.823}} & {\color{blue}\textbf{0.838}} & {\color{blue}\textbf{0.726}} & 0.640 & 0.337 & 0.898 & \multicolumn{1}{c|}{0.815} & {\color{blue}\textbf{0.7653}} \\
\hline
\multicolumn{1}{l|}{GraphBGS (semi-supervised)} & 0.837 & 0.942 & 0.700 & 0.743 & 0.405 & 0.558 & {\color{blue}\textbf{0.749}} & {\color{red}\textbf{0.966}} & \multicolumn{1}{c|}{0.729} & 0.7366 \\
\multicolumn{1}{l|}{GraphMOD-Net (semi-supervised, {\color{red}ours})} & 0.839 & {\color{blue}\textbf{0.955}} & 0.720 & {\color{red}\textbf{0.851}} & 0.554 & 0.521 & {\color{red}\textbf{0.770}} & {\color{blue}\textbf{0.942}} & \multicolumn{1}{c|}{0.682} & 0.7593 \\
\hline
\multicolumn{1}{l|}{BSUV-net (supervised)} & {\color{blue}\textbf{0.871}} & {\color{red}\textbf{0.969}} & 0.774 & 0.797 & {\color{red}\textbf{0.750}} & {\color{red}\textbf{0.680}} & 0.628 & 0.923 & \multicolumn{1}{c|}{{\color{red}\textbf{0.858}}} & {\color{red}\textbf{0.8056}} \\
\multicolumn{1}{l|}{FgSegNet v2 (supervised)} & 0.328 & 0.693 & 0.427 & 0.363 & 0.200 & 0.248 & 0.350 & 0.530 & \multicolumn{1}{c|}{0.604} & 0.4159 \\
\hline
\end{tabular}
\end{scriptsize}
\end{table*}

\begin{table*}
\centering
\caption{Comparison of the F-measure results over the videos of UCSD background subtraction dataset. The best and second best performing methods for each video are shown in {\color{red}\textbf{red}} and {\color{blue}\textbf{blue}}, respectively. DECOLOR and GRASTA are subspace learning methods.}
\label{tbl:UCSD_results}
\begin{scriptsize}
\begin{tabular}{l|cccccccc}
\hline
Sequence & \tiny{MoG} & \tiny{DECOLOR} & \tiny{ViBe} & \tiny{GRASTA} & \tiny{SuBSENSE} & \tiny{BSUV-Net} & \tiny{GraphBGS} & \tiny{GraphMOD} \\
\hline
Birds & 0.1427 & 0.1457 & 0.3354 & 0.1320 & 0.4832 & 0.2625 & \color{red}\textbf{0.7495} & \color{blue}\textbf{0.7391} \\
Boats & 0.0881 & 0.2179 & 0.1854 & 0.0678 & 0.4550 & 0.6621 & \color{blue}\textbf{0.7765} & \color{red}\textbf{0.8007} \\
Bottle & 0.1856 & 0.4765 & 0.4512 & 0.1159 & \color{blue}\textbf{0.6570} & 0.5039 & \color{red}\textbf{0.8741} & \color{red}\textbf{0.8741} \\
Chopper & 0.3237 & 0.6214 & 0.4930 & 0.0842 & 0.6723 & 0.3020 & \color{blue}\textbf{0.7766} & \color{red}\textbf{0.7844} \\
Cyclists & 0.0915 & 0.2224 & 0.1211 & 0.1243 & 0.1445 & 0.4138 & \color{blue}\textbf{0.7417} & \color{red}\textbf{0.7479} \\
Flock & 0.2706 & 0.2943 & 0.2306 & 0.1612 & 0.2492 & 0.0025 & \color{red}\textbf{0.5903} & \color{blue}\textbf{0.5871} \\
Freeway & 0.2622 & \color{blue}\textbf{0.5229} & 0.4002 & 0.0814 & \color{red}\textbf{0.5518} & 0.1185 & 0.3719 & 0.3516 \\
Hockey & 0.3867 & 0.3449 & 0.4195 & 0.3149 & 0.3611 & 0.6908 & \color{blue}\textbf{0.7664} & \color{red}\textbf{0.7804} \\
Jump & 0.2679 & 0.3135 & 0.2636 & 0.4175 & 0.2295 & \color{red}\textbf{0.8697} & \color{blue}\textbf{0.7734} & 0.7602 \\
Landing & 0.0335 & 0.0640 & 0.0433 & 0.0414 & 0.0026 & 0.0012 & \color{red}\textbf{0.2452} & \color{blue}\textbf{0.1840} \\
Ocean & 0.1113 & 0.1315 & 0.1648 & 0.1144 & 0.2533 & \color{blue}\textbf{0.5335} & \color{red}\textbf{0.8593} & \color{red}\textbf{0.8593} \\
Peds & 0.3731 & 0.7942 & 0.5257 & 0.4653 & 0.5154 & 0.6738 & \color{red}\textbf{0.8518} & \color{blue}\textbf{0.8512} \\
Skiing & 0.2038 & 0.3473 & 0.1441 & 0.0927 & 0.2482 & 0.0602 & \color{red}\textbf{0.5963} & \color{blue}\textbf{0.5953} \\
Surf & 0.0489 & 0.0647 & 0.0462 & 0.0523 & 0.0467 & 0 & \color{blue}\textbf{0.5851} & \color{red}\textbf{0.6139} \\
Surfers & 0.0542 & 0.1959 & 0.1189 & 0.0742 & 0.1393 & 0.4776 & \color{red}\textbf{0.6719} & \color{blue}\textbf{0.6611} \\
Trafic & 0.2188 & \color{blue}\textbf{0.2732} & 0.1445 & 0.0368 & 0.1165 & 0 & \color{red}\textbf{0.5722} & \color{red}\textbf{0.5722} \\
\hline
Overall & 0.1914 & 0.3144 & 0.2555 & 0.1485 & 0.3203 & 0.3483 & \color{red}\textbf{0.6751} & \color{blue}\textbf{0.6727} \\
\hline
\end{tabular}
\end{scriptsize}
\end{table*}

\subsection{Implementation Details}

The Cascade and Mask R-CNN were implemented using Pytorch and Detectron2 \cite{wu2019detectron2,zhang2020resnest}. The semi-supervised learning GCNN was implemented using TensorFlow \cite{abadi2016tensorflow}. The GCNN has a dropout rate of $0.5$ \cite{srivastava2014dropout}, an initial learning rate of $0.01$, a weight decay of $5\times 10^{-4}$, and a maximum number of $600$ epochs. We also implemented an early stopping with a window size of 10, \ie, the training is stopped when the validation loss does not decrease for 10 consecutive epochs.

For the comparison with GraphMOD-Net, the background subtraction algorithms MoG \cite{stauffer1999adaptive}, SuBSENSE \cite{st2014subsense}, and ViBe \cite{barnich2010vibe} for USCD were implemented with the BGSLibrary \cite{sobral2014bgs}; while GRASTA \cite{he2012incremental} and DECOLOR \cite{zhou2012moving} were implemented using the LRSLibrary \cite{lrslibrary2015}.

\section{Results and Discussion}
\label{sec:results_discussion}

GraphMOD-Net is compared, either in CDNet2014 or UCSD, with the following state-of-the-art algorithms for MOD: MoG \cite{stauffer1999adaptive}, GRASTA \cite{he2012incremental}, DECOLOR \cite{zhou2012moving}, Graph Cut Difference (GraphCutDiff) \cite{miron2015change}, ViBe \cite{barnich2010vibe}, SuBSENSE \cite{st2014subsense}, WisenetMD \cite{lee2019wisenetmd}, GraphBGS \cite{giraldo2020graphbgs}, BSUV-Net \cite{tezcan2020bsuv}, and FgSegNet v2 \cite{lim2019learning}. The results of FgSegNet v2 \cite{lim2019learning} for CDNet2014 are reported using unseen videos for the evaluation of the network, the performance comes from Tezcan \etal paper \cite{tezcan2020bsuv}.



Tables \ref{tbl:CDNet_results} and \ref{tbl:UCSD_results} show the comparison of GraphMOD-Net with several state-of-the-art methods in CDNet2014 and UCSD, respectively. The results of GraphMOD-Net and GraphBGS are computed with the best F-measures in each sequence from the experiments (both algorithms have the same instance segmentation method). For CDNet2014 in Table \ref{tbl:CDNet_results}, GraphMOD-Net shows the best results in the challenges baseline, dynamic background, PTZ, and shadow. The results also show that our algorithm has an overall increment in performance with respect to GraphBGS. For UCSD in Table \ref{tbl:UCSD_results}, GraphMOD-Net show either the best or the second-best results across almost all the videos of UCSD. In the overall performance, GraphMOD-Net comes second lagging slightly behind GraphBGS. We believe it is a consequence of increasing the representational power when solving the semi-supervised learning problem with GCNNs leading to a slightly worse generalization. Similarly, GraphMOD-Net outperforms subspace learning methods such as DECOLOR and GRASTA. Subspace learning methods generally have problems when dealing with moving camera sequences.

We can also notice the degradation of the deep learning method FgSegNet v2 compared to the original results \cite{lim2019learning} when tested on unseen videos in Table \ref{tbl:CDNet_results}.
Similarly, the performance of BSUV-Net drops when applied on UCSD, these results could suggest a weakness of deep learning methods in generalization for MOD (keep in mind that BSUV-Net uses the output of a fully CNN trained in ADE20K dataset \cite{zhou2017scene} as input of the network). In Table \ref{tbl:UCSD_results}, we evaluated the BSUV-Net model provided by the authors of \cite{tezcan2020bsuv} that was trained in the CDNet2014. In our case, we trained GraphMOD-Net in a small subset of nodes of CDNet2014 including the challenges baseline, dynamic background, shadow, and PTZ. This suggests that GraphMOD-Net generalizes better than BSUV-Net when applied on a different dataset from which it was trained. For the sake of clarification, we did not use the ground truth of UCSD in the training procedure.


\begin{figure*}
    \centering
    \includegraphics[width=0.815\textwidth]{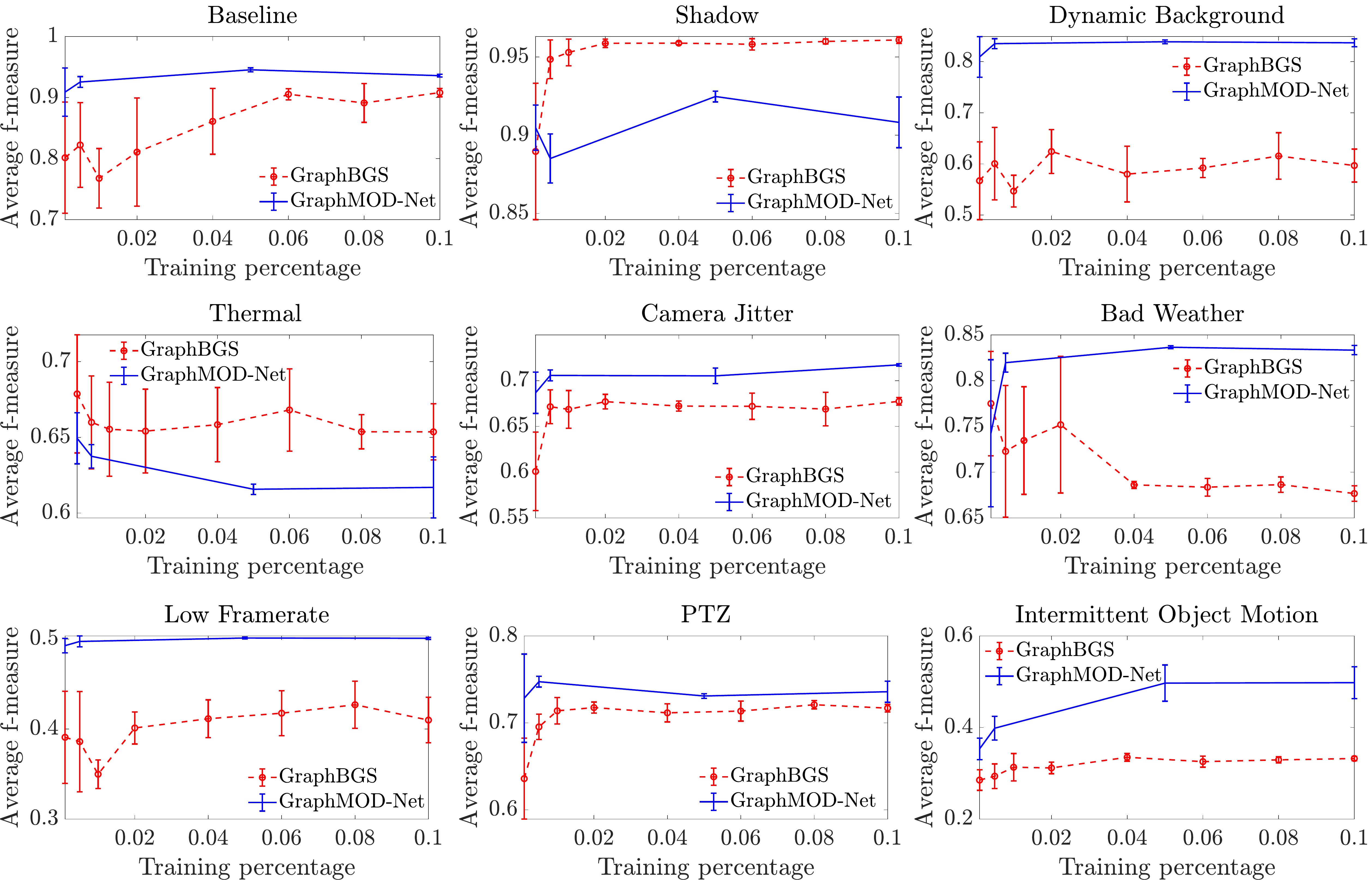}
    \caption{Average F-measure vs training percentage in the nine challenges of CDNet2014, using GraphBGS, and our GraphMOD-Net. Each point in the plots is a Monte Carlo cross-validation experiment with 3 repetitions.}
    \label{fig:experimental_results}
\end{figure*}

Figure \ref{fig:experimental_results} depicts the results of our GraphMOD-Net compared with the previous semi-supervised MOD method GraphBGS \cite{giraldo2020graphbgs}. The results of the challenges turbulence and night videos are not displayed since our Mask R-CNN fails to detect several instances in these videos. As a result, it is not possible to detect the moving objects in some sequences for the GCNN. GraphMOD-Net is better than GraphBGS in the challenges of bad weather, baseline, camera jitter, dynamic background, intermittent object motion, low frame rate, and PTZ; while GraphBGS shows better results than GraphMOD-Net in the challenges shadow and thermal. GraphMOD-Net benefits from the higher modeling capacity of GCNNs by improving upon the GraphBGS as shown in Tables \ref{tbl:CDNet_results}, \ref{tbl:UCSD_results}, and in Figure \ref{fig:experimental_results}.

\begin{table*}
\centering
\caption{Some visual results of GraphMOD-Net in CDNet2014 compared with state-of-the-art algorithms, from left to right: original images, ground-truth images, BSUV-Net \cite{tezcan2020bsuv}, GraphBGS \cite{giraldo2020graphbgs}, and the proposed GraphMOD-Net.}
\label{tbl:visual_results}
\makebox[\linewidth]{
\scalebox{0.88}{
\begin{tabular}{|lccccc|}
\hline 
\textbf{Categories} & Original & Ground Truth  & BSUV-Net & GraphBGS & GraphMOD-Net \\
\hline 
\hline 
\begin{tabular}[l]{@{}l@{}}Bad Weather\\ wetSnow\\ in000500\end{tabular} & \parbox[c][16mm][c]{18mm}{\includegraphics[height=13.5mm]{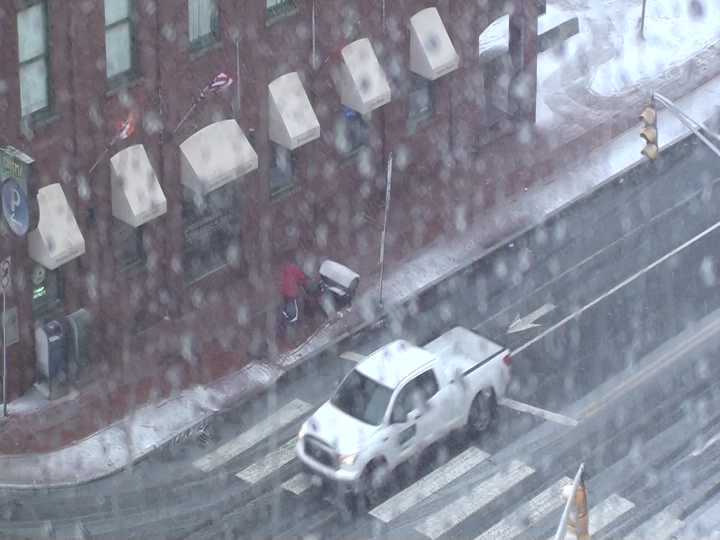}} & \parbox[c][16mm][c]{18mm}{\includegraphics[height=13.5mm]{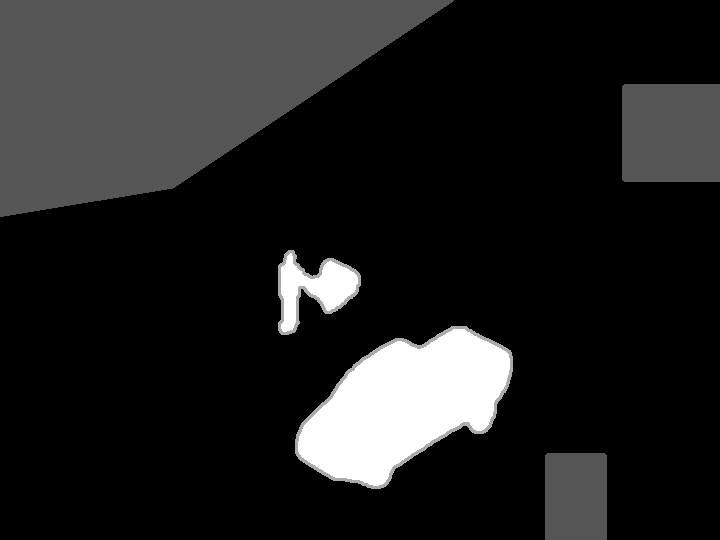}} & \parbox[c][16mm][c]{18mm}{\includegraphics[height=13.5mm]{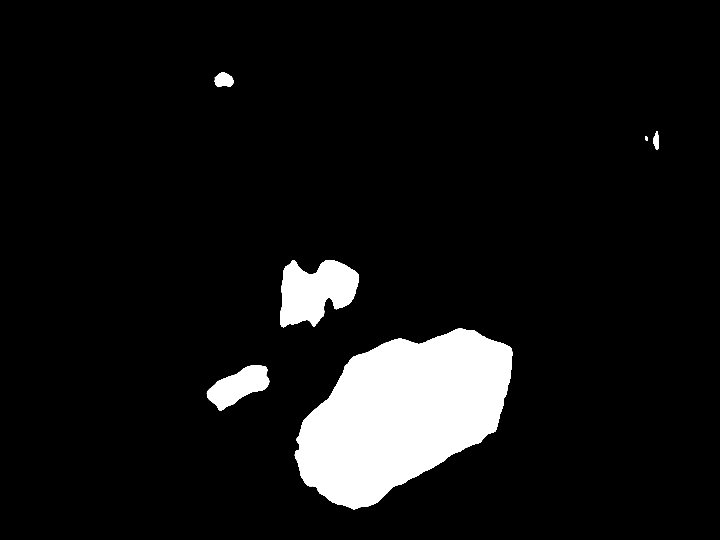}} & \parbox[c][16mm][c]{18mm}{\includegraphics[height=13.5mm]{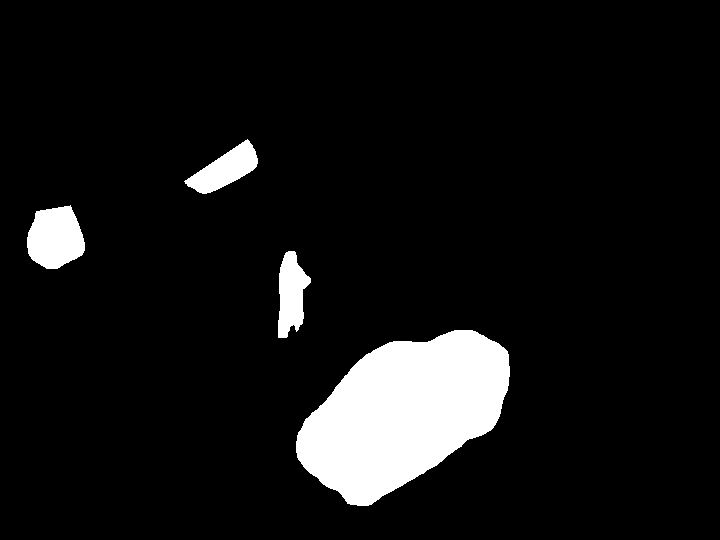}} &  \parbox[c][16mm][c]{18.5mm}{\includegraphics[height=13.5mm]{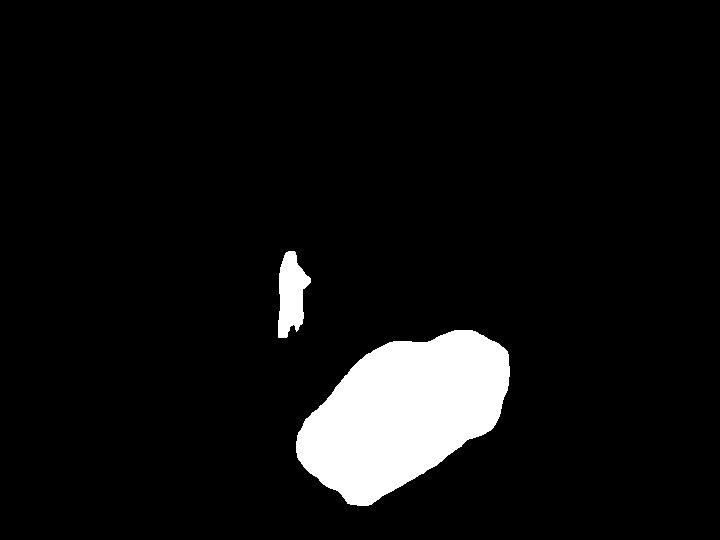}} \\ 
\hline  
\begin{tabular}[l]{@{}l@{}}Baseline\\ Pedestrians\\ in000496\end{tabular} & \parbox[c][15mm][c]{18mm}{\includegraphics[height=12mm]{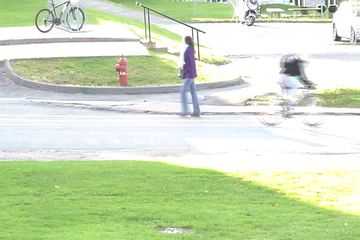}} & \parbox[c][15mm][c]{18mm}{\includegraphics[height=12mm]{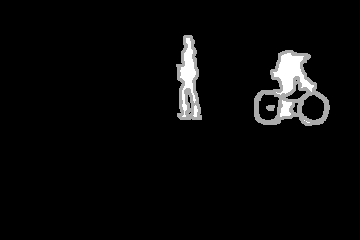}} & \parbox[c][15mm][c]{18mm}{\includegraphics[height=12mm]{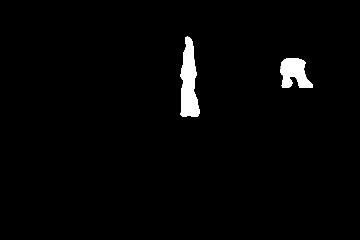}} & \parbox[c][15mm][c]{18mm}{\includegraphics[height=12mm]{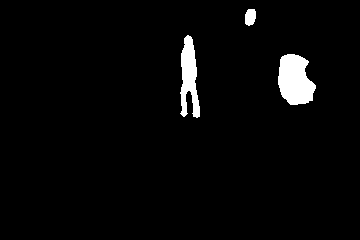}} &  \parbox[c][15mm][c]{18.5mm}{\includegraphics[height=12mm]{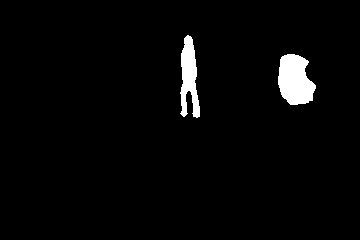}} \\ 
\hline  
\begin{tabular}[l]{@{}l@{}}Camera Jitter\\ Boulevard\\ in000915\end{tabular} & \parbox[c][15mm][c]{18mm}{\includegraphics[height=12.2mm]{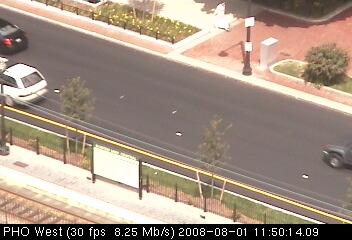}} & \parbox[c][15mm][c]{18mm}{\includegraphics[height=12.2mm]{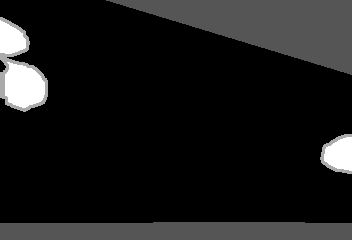}} & \parbox[c][15mm][c]{18mm}{\includegraphics[height=12.2mm]{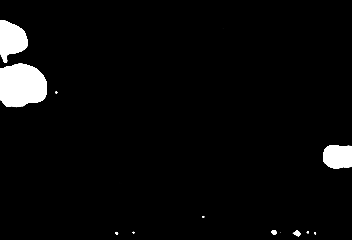}} & \parbox[c][15mm][c]{18mm}{\includegraphics[height=12.2mm]{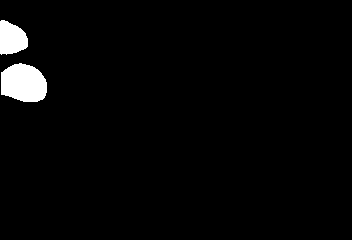}} &  \parbox[c][15mm][c]{18.5mm}{\includegraphics[height=12.2mm]{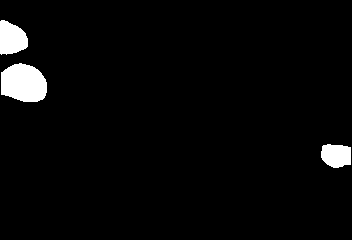}} \\ 
\hline  
\begin{tabular}[l]{@{}l@{}}Dynamic Back.\\ Fall\\ in002789\end{tabular} & \parbox[c][15mm][c]{18mm}{\includegraphics[height=12mm]{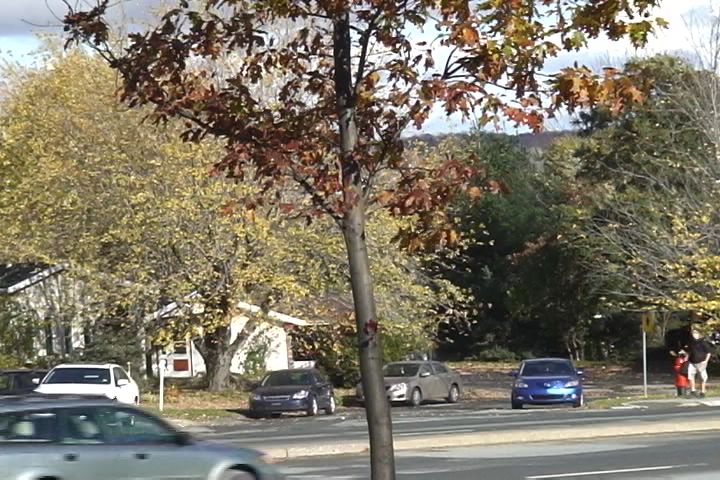}} & \parbox[c][15mm][c]{18mm}{\includegraphics[height=12mm]{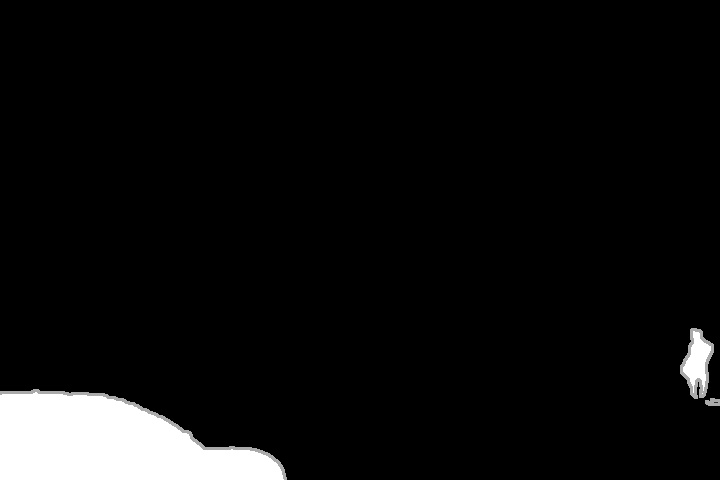}} & \parbox[c][15mm][c]{18mm}{\includegraphics[height=12mm]{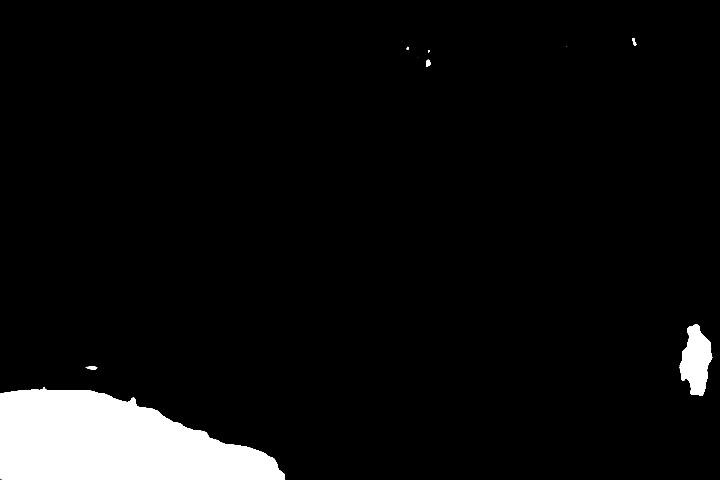}} & \parbox[c][15mm][c]{18mm}{\includegraphics[height=12mm]{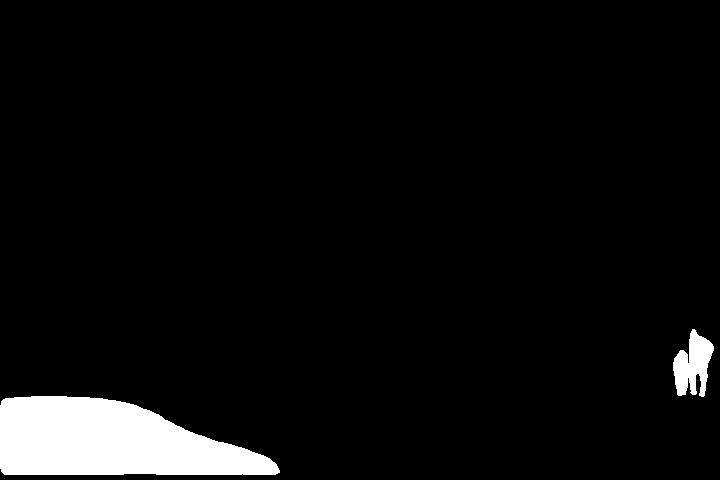}} &  \parbox[c][15mm][c]{18.5mm}{\includegraphics[height=12mm]{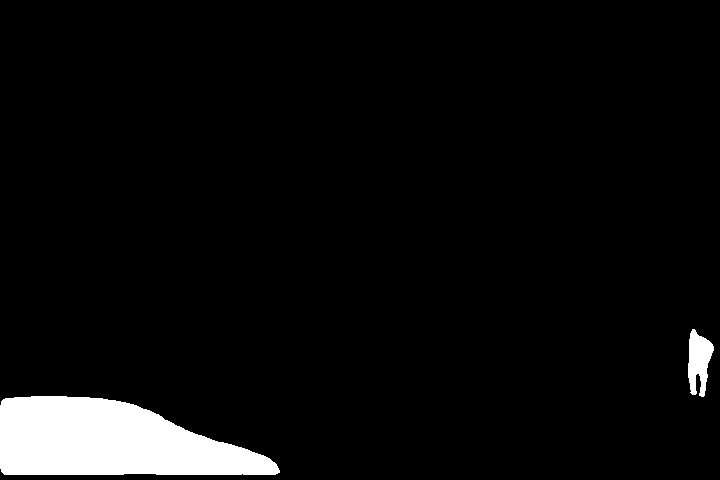}} \\ 
\hline  
\begin{tabular}[l]{@{}l@{}}PTZ\\ Two Position.\\ in002789\end{tabular} & \parbox[c][14mm][c]{18mm}{\includegraphics[height=10.8mm]{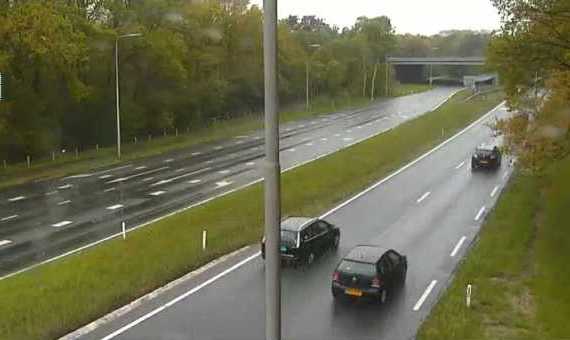}} & \parbox[c][14mm][c]{18mm}{\includegraphics[height=10.8mm]{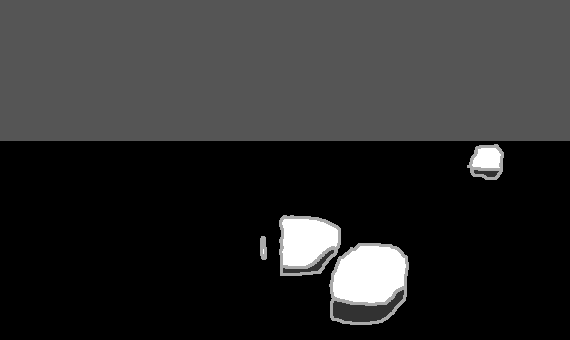}} & \parbox[c][14mm][c]{18mm}{\includegraphics[height=10.8mm]{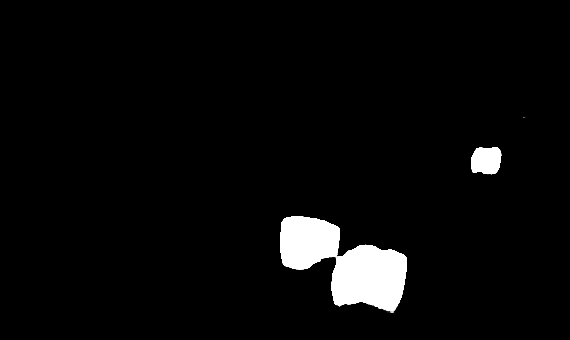}} & \parbox[c][14mm][c]{18mm}{\includegraphics[height=10.8mm]{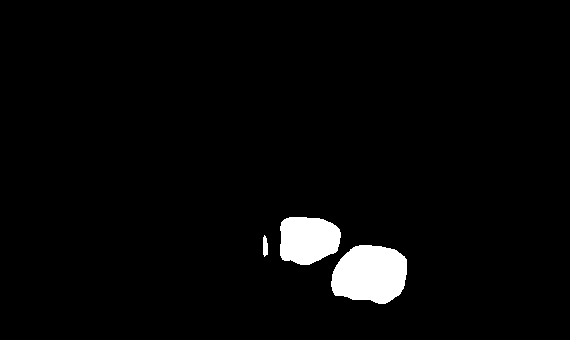}} &  \parbox[c][14mm][c]{18.5mm}{\includegraphics[height=10.8mm]{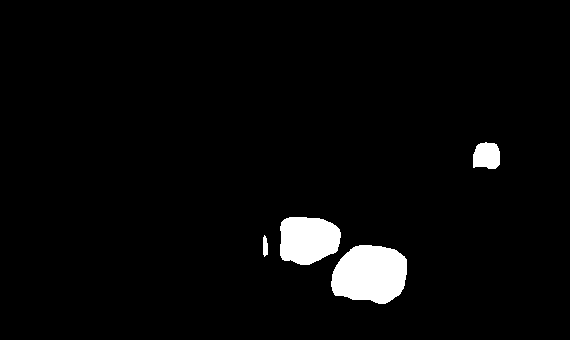}} \\ 
\hline  
\begin{tabular}[l]{@{}l@{}}Low-F Rate\\ TramCross\\ in000607\end{tabular} & \parbox[c][13mm][c]{18mm}{\includegraphics[height=9.8mm]{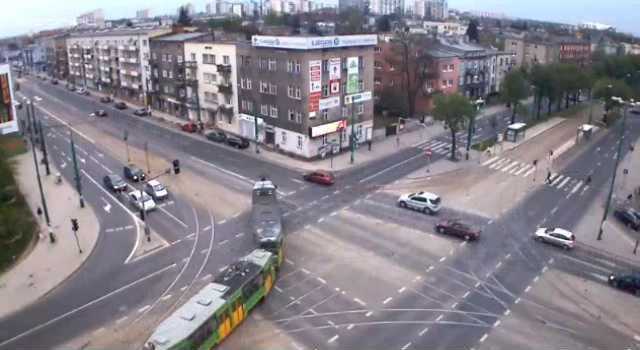}} & \parbox[c][13mm][c]{18mm}{\includegraphics[height=9.8mm]{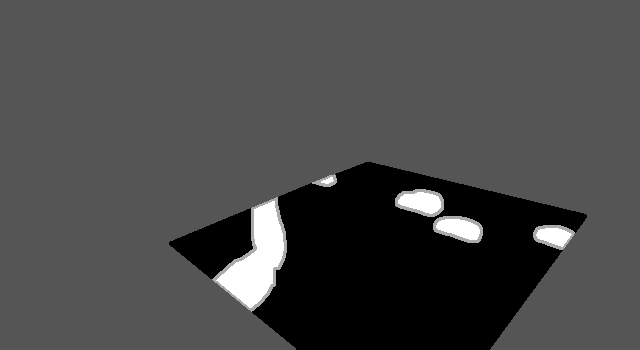}} & \parbox[c][13mm][c]{18mm}{\includegraphics[height=9.8mm]{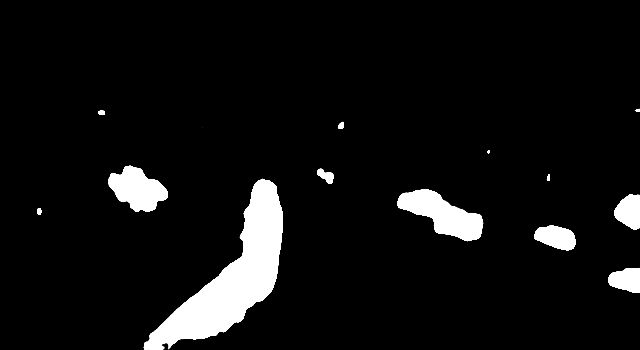}} & \parbox[c][13mm][c]{18mm}{\includegraphics[height=9.8mm]{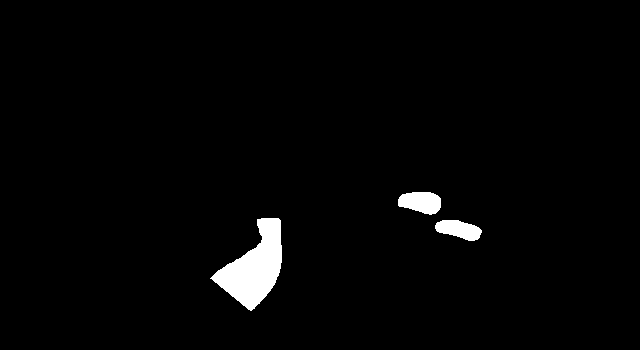}} &  \parbox[c][13mm][c]{18.5mm}{\includegraphics[height=9.8mm]{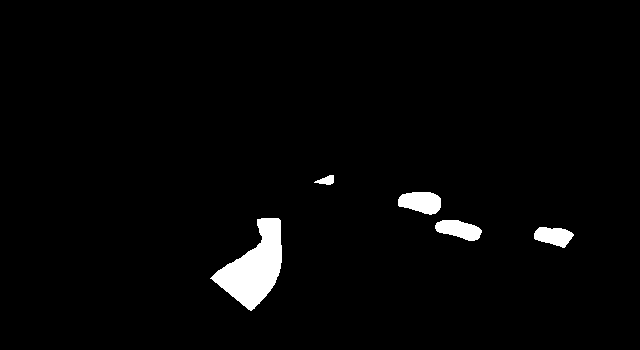}} \\ 
\hline  
\end{tabular}
}
}
\end{table*}

\begin{table*}
\centering
\caption{Some visual results of GraphMOD-Net in UCSD compared with state-of-the-art algorithms, from left to right: original images, ground-truth images, DECOLOR \cite{zhou2012moving}, SuBSENSE \cite{st2014subsense}, and the proposed GraphMOD-Net.}
\label{tbl:visual_results_UCSD}
\makebox[\linewidth]{
\scalebox{0.88}{
\begin{tabular}{|lccccc|}
\hline 
\textbf{Videos} & Original & Ground Truth & DECOLOR & SuBSENSE & GraphMOD-Net \\
\hline 
\hline 
\begin{tabular}[l]{@{}l@{}}Birds\\ frame\_4\end{tabular} & \parbox[c][16mm][c]{18.5mm}{\includegraphics[height=12mm]{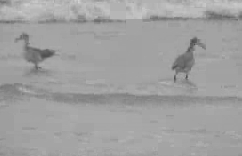}} & \parbox[c][16mm][c]{18.5mm}{\includegraphics[height=12mm]{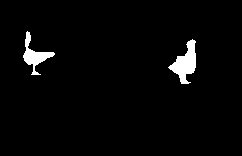}} & \parbox[c][16mm][c]{18.5mm}{\includegraphics[height=12mm]{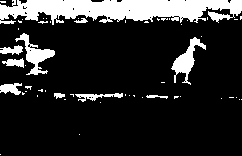}} & \parbox[c][16mm][c]{18.5mm}{\includegraphics[height=12mm]{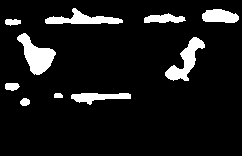}} &  \parbox[c][16mm][c]{18.5mm}{\includegraphics[height=12mm]{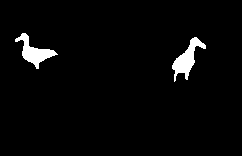}} \\ 
\hline 
\begin{tabular}[l]{@{}l@{}}Boats\\ frame\_12\end{tabular} & \parbox[c][16mm][c]{18.5mm}{\includegraphics[height=12.1mm]{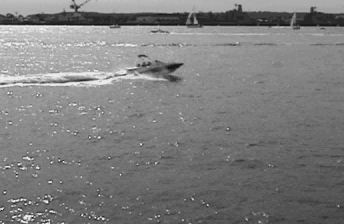}} & \parbox[c][16mm][c]{18.5mm}{\includegraphics[height=12.1mm]{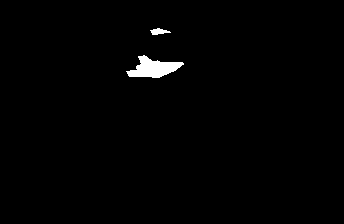}} & \parbox[c][16mm][c]{18.5mm}{\includegraphics[height=12.1mm]{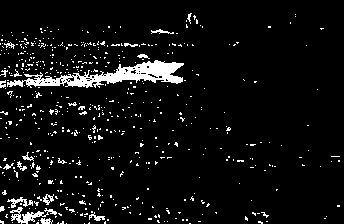}} & \parbox[c][16mm][c]{18.5mm}{\includegraphics[height=12.1mm]{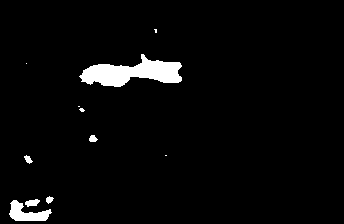}} &  \parbox[c][16mm][c]{18.5mm}{\includegraphics[height=12.1mm]{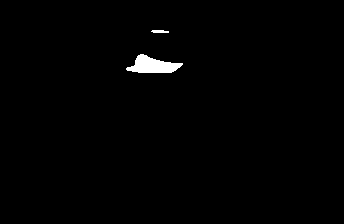}} \\ 
\hline 
\begin{tabular}[l]{@{}l@{}}Bottle\\ frame\_14\end{tabular} & \parbox[c][16.5mm][c]{18.5mm}{\includegraphics[height=13.6mm]{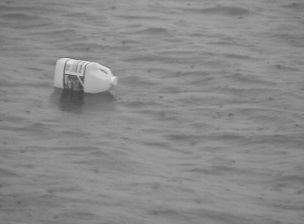}} & \parbox[c][16.5mm][c]{18.5mm}{\includegraphics[height=13.6mm]{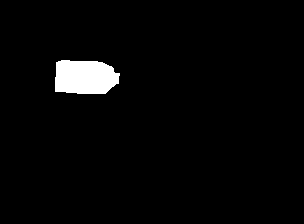}} & \parbox[c][16.5mm][c]{18.5mm}{\includegraphics[height=13.6mm]{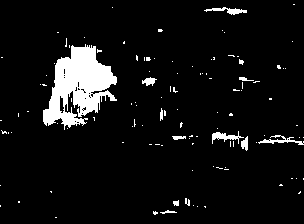}} & \parbox[c][16.5mm][c]{18.5mm}{\includegraphics[height=13.6mm]{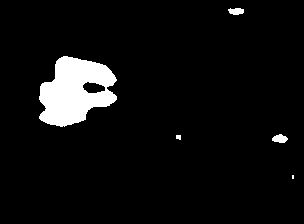}} &  \parbox[c][16.5mm][c]{18.5mm}{\includegraphics[height=13.6mm]{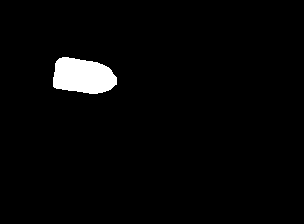}} \\ 
\hline
\begin{tabular}[l]{@{}l@{}}Chopper\\ frame\_47\end{tabular} & \parbox[c][16mm][c]{18.5mm}{\includegraphics[height=12mm]{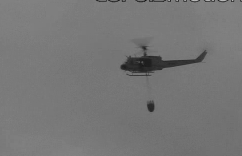}} & \parbox[c][16mm][c]{18.5mm}{\includegraphics[height=12mm]{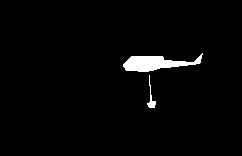}} & \parbox[c][16mm][c]{18.5mm}{\includegraphics[height=12mm]{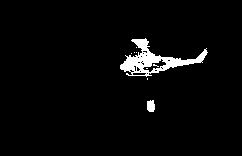}} & \parbox[c][16mm][c]{18.5mm}{\includegraphics[height=12mm]{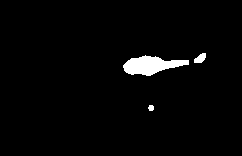}} &  \parbox[c][16mm][c]{18.5mm}{\includegraphics[height=12mm]{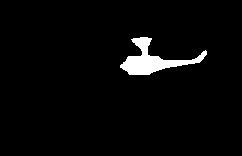}} \\ 
\hline
\begin{tabular}[l]{@{}l@{}}Cyclists\\ frame\_9\end{tabular} & \parbox[c][14mm][c]{18.5mm}{\includegraphics[height=10.3mm]{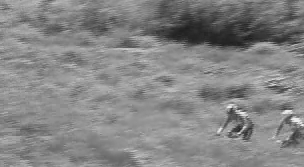}} & \parbox[c][14mm][c]{18.5mm}{\includegraphics[height=10.3mm]{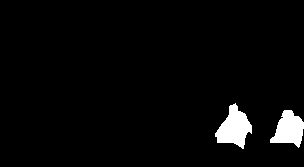}} & \parbox[c][14mm][c]{18.5mm}{\includegraphics[height=10.3mm]{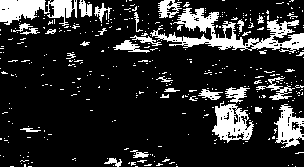}} & \parbox[c][14mm][c]{18.5mm}{\includegraphics[height=10.3mm]{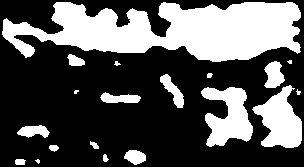}} &  \parbox[c][14mm][c]{18.5mm}{\includegraphics[height=10.3mm]{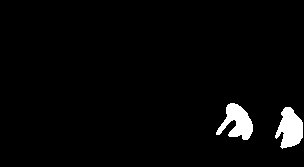}} \\ 
\hline
\begin{tabular}[l]{@{}l@{}}Peds\\ frame\_25\end{tabular} & \parbox[c][16mm][c]{18.5mm}{\includegraphics[height=12.2mm]{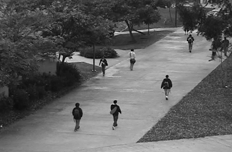}} & \parbox[c][16mm][c]{18.5mm}{\includegraphics[height=12.2mm]{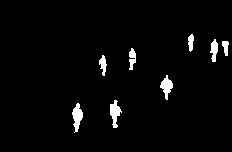}} & \parbox[c][16mm][c]{18.5mm}{\includegraphics[height=12.2mm]{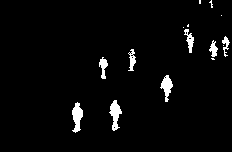}} & \parbox[c][16mm][c]{18.5mm}{\includegraphics[height=12.2mm]{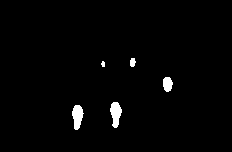}} &  \parbox[c][16mm][c]{18.5mm}{\includegraphics[height=12.2mm]{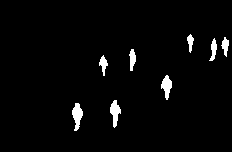}} \\ 
\hline  
\end{tabular}
}
}
\end{table*}

Table \ref{tbl:visual_results} shows some qualitative results of GraphMOD-Net compared with BSUV-Net \cite{tezcan2020bsuv}, and GraphBGS \cite{giraldo2020graphbgs} for CDNet2014. In the same way, Table \ref{tbl:visual_results_UCSD} shows visual results of GraphMOD-Net on UCSD compared with DECOLOR and SuBSENSE. For a fair comparison, the results in Table \ref{tbl:visual_results} for GraphBGS and GraphMOD-Net are computed with the same training percentage on each sequence. The visual results of GraphMOD-Net indicate that GCNNs are better than previous unsupervised and supervised algorithms used for MOD in some challenges. The improvement of GraphMOD-Net with respect to GraphBGS is because of the modeling capacity of GCNNs compared with previous semi-supervised learning methods, where a smooth prior assumption is required.

\section{Conclusions}
\label{sec:conclusions}

This paper introduces a method for MOD based on GCNN. The pipeline of the method involves a Cascade Mask R-CNN or a Mask R-CNN to get the object instances from the videos; a graph construction with k-nearest neighbors, where nodes are associated to object instances, and represented with feature vector encompassing optical flow, intensity and texture descriptors. We use a compact GCNN model trained in a semi-supervised mode.
Our algorithm outperforms previous state-of-the-art unsupervised, semi-supervised, and supervised learning algorithms for MOD in several challenges of the CDNet2014 and UCSD datasets.
In the future, we will further study the implications of the architecture of GCNNs in the problem of MOD. In the same way, we will analyze some theoretical aspects in the training of the GCNNs with focus in MOD. And finally, we will study a inductive framework \cite{hamilton2017inductive} for GraphMOD-Net in the context of MOD.

{\small
\bibliographystyle{ieee_fullname}
\bibliography{egbib}
}

\end{document}